\documentclass{article}

\usepackage[preprint]{neurips_2025}

\usepackage[utf8]{inputenc} 
\usepackage[T1]{fontenc}    
\usepackage{hyperref}       
\usepackage{url}            
\usepackage{booktabs}       
\usepackage{amsfonts}       
\usepackage{nicefrac}       
\usepackage{microtype}      
\usepackage{xcolor}         
\usepackage{pifont}
\usepackage{enumitem}
\usepackage{booktabs}
\usepackage{multirow}
\usepackage{amsmath}
\usepackage{xspace}
\usepackage{xcolor}
\usepackage{algorithm}
\usepackage{algorithmic}
\usepackage{wrapfig}
\usepackage{graphicx}

\usepackage{xcolor}
\definecolor{darkgreen}{RGB}{0,150,0}
\definecolor{darkred}{RGB}{150,0,0}

\usepackage{hyperref}
\usepackage{amssymb}
\usepackage{orcidlink}
\usepackage[table]{xcolor}

\definecolor{lightgray}{RGB}{238,238,238}

\newcommand{\sys}{\textsc{Sieve}\xspace}

\title{Improving Visual Reasoning with Iterative Evidence Refinement}

\author{
Zeru Shi\thanks{Equal contribution} \quad
Kai Mei\footnotemark[1] \quad
Yihao Quan \quad
Dimitris N. Metaxas \quad
Ruixiang Tang\thanks{Corresponding author} \\ \\
Department of Computer Science \\
\textbf{Rutgers University}
}

\begin{document}

\maketitle

\begin{abstract}

Vision–language models (VLMs) are increasingly capable of reasoning over images, but robust visual reasoning often requires re-grounding intermediate steps in the underlying visual evidence. Recent approaches typically rely on external image operations such as zooming or cropping to re-access fine-grained details during inference, which requires additional image re-encoding and can disrupt the reasoning trajectory. We argue that VLMs already provide strong internal signals for identifying and reusing visual evidence, and that these signals can be directly leveraged to support image-grounded reasoning. Motivated by this insight, we propose an end-to-end self-revisit framework, \textbf{\sys}, that trains models to re-engage image evidence through internal representations. \sys automatically extracts embeddings of salient image regions and injects them into the reasoning chain when additional grounding is needed, enabling later steps to condition on relevant visual cues without external tool calls or re-encoding. We use reinforcement learning to teach the model when to trigger visual revisiting and which region embeddings to retrieve and insert during the reasoning process. Experiments on multiple visual reasoning benchmarks, together with perception, reasoning, and hallucination evaluations, show that \sys yields consistent gains, improving performance by 8\% on average across several benchmarks.

\end{abstract}
    
\section{Introduction}
\label{sec:intro}

Vision–Language Models (VLMs) have demonstrated strong reasoning performance on multimodal question answering, often supported by long chain-of-thought (CoT) reasoning~\cite{team2025kimi, guo2025seed1, bai2025qwen2, team2025kimi1}. However, their reasoning pipeline remains largely text-centric. In a standard VLM inference pipeline, the image is encoded into a fixed set of visual tokens that serve as static context, while reasoning unfolds autoregressively in text. As generation proceeds, the model’s conditioning gradually shifts toward the growing history of generated text tokens, reducing the relative influence of visual evidence~\cite{li2025hidden,fang2025grounding}. Consequently, the model rarely revisits the image in a targeted, step-dependent way, and visual information is often underutilized in long-horizon reasoning.


To address this limitation, recent work has begun to explicitly integrate visual evidence into the reasoning trajectory, drawing inspiration from human cognition~\cite{shao2024visual}. Models repeatedly consult the image during CoT to improve grounded reasoning. After the release of OpenAI’s o3 model~\cite{o3}, a common implementation has operationalized this idea through external visual operations such as zooming and cropping, generating sub-images for targeted inspection~\cite{huang2025sam,bai2025univg}. Beyond predefined operations, other approaches allow VLMs to generate executable code for more flexible, programmatic image manipulation~\cite{lee2025interactive,mallis2025cad}.

\begin{figure}[t]
\centering
\includegraphics[width=0.7\linewidth]{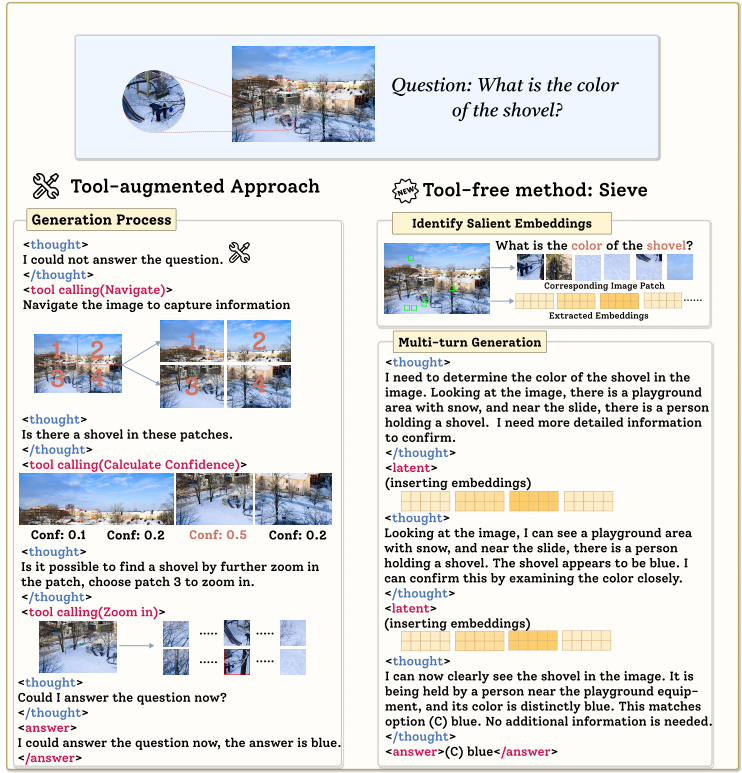}
\caption{This figure compares tool-augmented methods with \sys. The left shows tool-based reasoning, where external tools are invoked for additional visual information. The right shows \sys, which directly retrieves and injects key region embeddings into the reasoning process.}
\label{fig:heat1}
\vspace{-20pt}
\end{figure}

Despite their effectiveness, existing methods are constrained by their reliance on external tools or agents, which introduces two key drawbacks. First, existing methods often disrupt the continuity of the reasoning chain during multi-turn interaction. Specifically, these approaches typically rely on an external module to generate a new view of the image, which is then appended to the original image input rather than being directly integrated into the CoT reasoning process. As a result, the extracted visual view is not inserted into the corresponding positions of the CoT; instead, it is appended to the input as additional images before the input and current output text. Second, enabling the model to repeatedly invoke external tools requires constructing a large amount of training data and designing complex training pipelines for the VLM to learn such capabilities.

In this paper, we challenge the prevailing paradigm and ask a simple question: \textit{do we truly need to generate new image views through external operations} to let model revist the image information during inference? Our hypothesis is that the original visual embeddings already contain sufficient information for grounded reasoning, and that the main bottleneck is the model’s limited ability to selectively reuse relevant visual evidence as generation unfolds. Instead of cropping, zooming, and re-encoding additional images, we propose to directly extract task-relevant visual embeddings and insert them into the reasoning chain. To validate this idea, we conduct a preliminary analysis by manually identifying salient regions and directly injecting their visual embeddings into intermediate reasoning. This simple intervention consistently improves performance on the V* benchmark, yielding a 3\% accuracy gain without any additional training.

Inspired by this observation, we introduce \textbf{\sys}, a framework that enables VLMs to revisit visual evidence without external tools or agentic image operations. As shown in \autoref{fig:heat1}, when the model signals a need for additional visual grounding during inference, \sys retrieves embeddings for the relevant image regions and inserts them into the current reasoning chain, rather than localizing regions with an external tool and re-encoding new views. By reusing already encoded region features, \sys preserves access to fine-grained, localized visual cues for grounded multi-step reasoning while avoiding redundant vision re-encoding. We further develop a visually grounded RL training pipeline that enables the model to learn when and how to effectively retrieve and insert region embeddings. This RL training is highly data-efficient: \sys requires only a small dataset (approximately 1.5k samples) to acquire the capability. We evaluate \sys on Qwen3-based VLMs (4B and 8B)~\citep{yang2025qwen3} across multiple benchmarks. Results show consistent improvements over inference time tool-augmented baselines, indicating that much of the benefit typically attributed to explicit visual re-inspection can be achieved through retrieval and reuse of the embeddings of visual evidence. Our contributions are:
\begin{itemize}[leftmargin=*]
    \item We propose \textbf{\sys}, a framework that lets VLMs revisit fine-grained visual evidence by retrieving and reinserting region-level visual embeddings from the original encoding, avoiding external crop/zoom tools and any re-encoding.
    \item We design a saliency-based mechanism to identify the embeddings of visual evidence that could be critical to reasoning semantics. Building on this, we develop a visually grounded RL training pipeline that teaches the model when to retrieve these visual embeddings during reasoning. 
    \item We validate \sys across multiple model scales and benchmarks, where we show that using only a small training set (approximately 1.5k samples), \sys can demonstrate consistent gains in grounded multimodal performance (up to 8\% on average across several benchmarks).
\end{itemize} 
\section{Related Work}

\subsection{Tool-augmented Multi-modal Reasoning}
Recent work extends VLM inference beyond a single visual pass by introducing explicit visual re-query mechanisms during reasoning. A prominent line equips models with visual tools (e.g., cropping, zooming, object detection) that produce targeted observations and feed them back as additional inputs~\cite{huang2025visualtoolagent, su2025openthinkimg, fan2025grit, chen2025sifthinker, huvisual}. Later approaches learn tool-use policies with reinforcement learning, rewarding effective re-query trajectories and verification behaviors, often adopting coarse-to-fine strategies that start from a global view and refine under higher-resolution observations~\cite{zheng2025deepeyes, zhong2025omni}. Related methods further encourage pixel-space exploration, either via instruction tuning to broaden search coverage~\cite{su2025pixel} or RL to strengthen perceptual competence and steer attention to task-relevant regions~\cite{yu2025perception}. Despite strong performance, these systems typically incur inference-time overhead due to repeated view generation and re-encoding.

\subsection{Multi-modal Reasoning in Latent Space}
In parallel, several lines reduce reliance on explicit image operations by operating in representation space or by selectively using visual tokens. Generation-based methods synthesize auxiliary visual representations to support inference~\cite{xu2025visual, chern2025thinking, li2025imagine}. More recently, latent thinking-with-images paradigms introduce learnable latent visual tokens and embedding-level manipulation to internalize certain visual operations and enable mode switching during inference~\cite{li2025latent, zhang2025deepsketcher, yang2025machine}. A complementary efficiency literature observes that dense patch sequences are redundant and studies how to select, prune, or merge visual tokens while preserving representational fidelity~\cite{chen2024spatialvlm, cao2024madtp, wang2025sota, bolya2022token, zeng2025glimpse, zhang2024area, huang2024ivtp, li2025regate, hu2025tokenflex, yu2025introducing, song2024less}. These approaches share a common premise: they construct reasoning processes within a latent visual space and train models to perform reasoning in that space. However, this paradigm requires substantial effort to enable the model to learn reasoning over newly introduced visual latents that differ from the native textual representation space. Motivated by this limitation, we propose a framework that directly leverages the embeddings of images within the textual reasoning space, rather than constructing and training reasoning in a separate latent space. 

\section{Methodology}
\label{sec:method}

\begin{wrapfigure}{r}{0.35\textwidth}
    \centering
    \includegraphics[width=\linewidth]{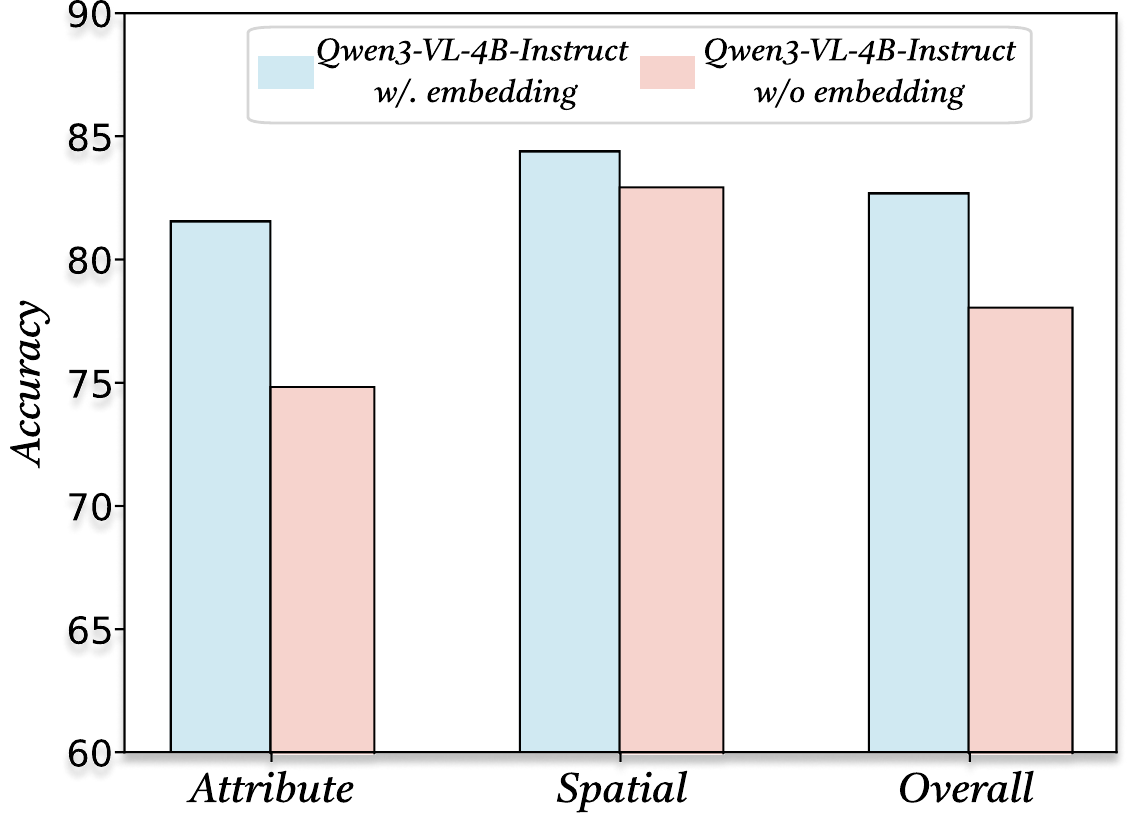}
    \caption{Performance with and without region embeddings.}
    \label{fig:preexp}
    \vspace{-20pt}
\end{wrapfigure}

Our core hypothesis is that the visual embeddings produced by a VLM already encode sufficient information for complex visual reasoning, provided the model can access the appropriate localized evidence at the right moment. To validate this hypothesis, we conduct a controlled study on the V* dataset examining whether region-level visual features can directly enhance multimodal inference. Concretely, we manually identify task-relevant regions, extract their corresponding embeddings, and augment inference by inserting these region embeddings with the model's original visual embeddings, all without any additional training. We evaluate this intervention against the standard setting that relies solely on global image. As shown in \autoref{fig:preexp}, region-augmented inference yields a 3\% improvement, confirming that localized embedding evidence provides a compact yet effective signal that the model can immediately leverage. This finding aligns with prior work demonstrating that LLM hidden states encode rich semantic structure~\citep{skean2025layer, schiekiera2026associations, liu2024fantastic}.

Motivated by this observation, we introduce \textbf{\sys}, a framework that treats task-relevant region embeddings as reusable visual evidence and learns to incorporate them within RL policy optimization. Specifically, \sys (i) extracts and caches region embeddings as compact evidence units and (ii) jointly optimizes how such evidence is selected and integrated into the model's reasoning process throughout reinforcement learning. Section~\ref{sec:workflow} presents an overview of the training pipeline. Section~\ref{sec:training_data} details our automatic evidence discovery procedure. Section~\ref{sec:embedding_rl} introduces a visually grounded RL formulation that trains the model to retrieve and insert cached region embeddings on demand, enabling systematic evidence-aware reasoning beyond global visual representations.

\begin{figure}[t]
    \centering
    \includegraphics[width=\linewidth]{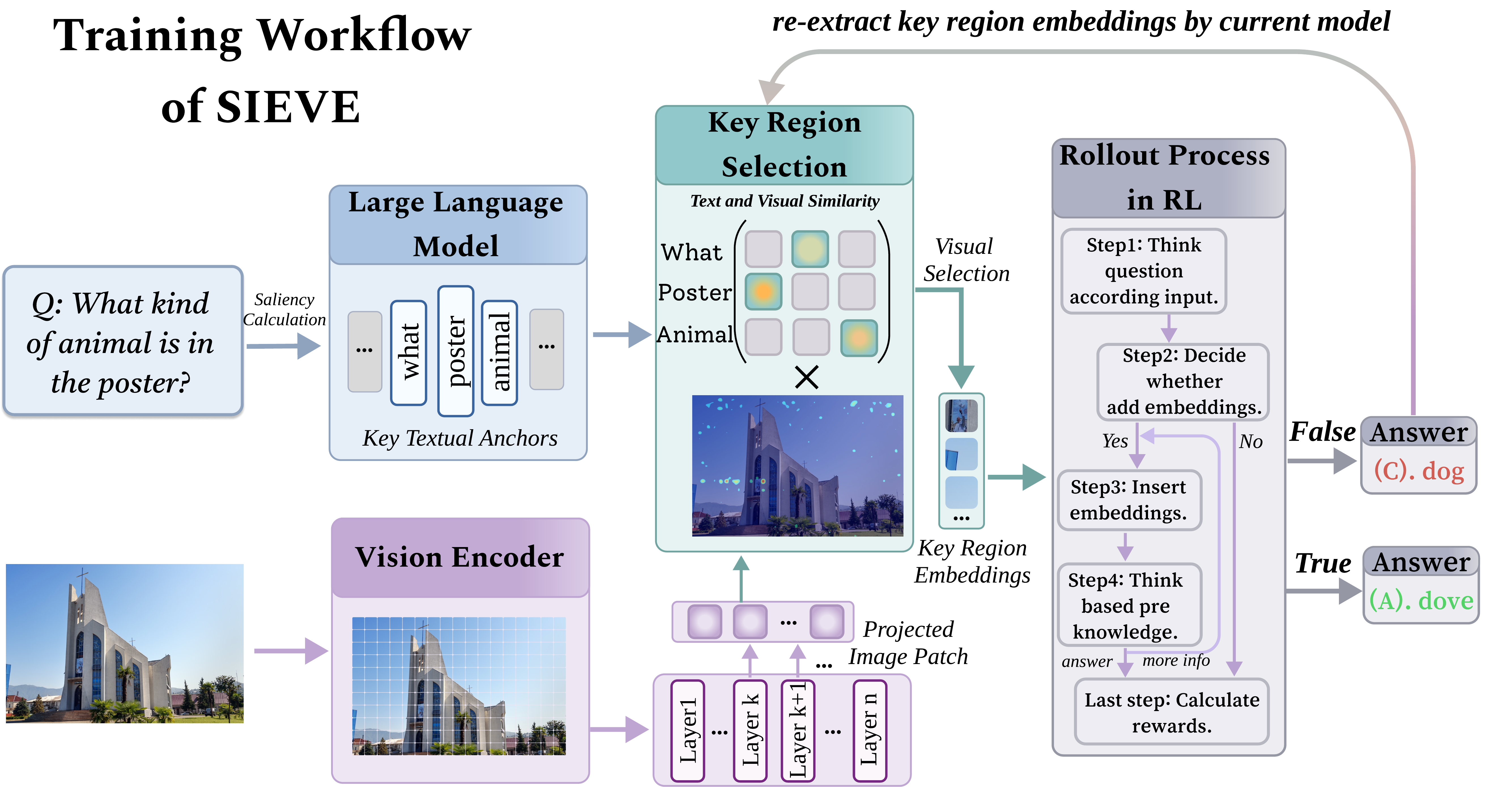}
    \caption{Training workflow of SIEVE. For each question, the embeddings of image patches aligned with key textual anchors are extracted and cached as visual evidence. During RL rollouts, the policy learns when to insert this evidence into the reasoning stream, with rewards computed from the final answer. Embeddings of visual evidence are periodically re-extracted using the updated model to keep the evidence aligned with the evolving policy.}
    \label{fig:workflow}
    \vspace{-10pt}
\end{figure}
\subsection{Overview of \sys}
\label{sec:workflow}

\autoref{fig:workflow} illustrates the training workflow of \sys. Prior to training-time rollouts, we construct embeddings of visually salient regions through a two-stage process. On the textual side, we compute token-level saliency scores to identify the tokens that exert the greatest influence on subsequent generation. These salient tokens serve as semantic queries. On the visual side, we compute cross-modal similarity between each query token and the image patch representations to localize the most relevant spatial regions.

\begin{algorithm}[t]
\caption{Self-Guided Visual Evidence Discovery}
\label{alg:evidence_discovery}
\small
\begin{algorithmic}
\REQUIRE Multimodal model $\mathcal{M}$; image $I$; token embeddings $\{\mathbf{h}_i\}_{i=1}^{L}$; $u$, $v$ row and column coordinates of the patch.
\ENSURE Evidence snapshot set $\mathcal{E}$

\STATE Run $\mathcal{M}(I,\{\mathbf{h}_i\})$ to obtain logits $\mathbf{z}_L$ and hidden states $\{\mathbf{H}^{(\ell)}\}$.
\STATE $\hat v=\arg\max_{v\in\mathcal{V}}\mathbf{z}_L[v],\; s=\mathbf{z}_L[\hat v]$ \hfill{\footnotesize // choose prediction target}

\STATE $\mathrm{Sal}(i)=\|\nabla_{\mathbf{h}_i}s\odot \mathbf{h}_i\|_2,\;
\mathcal{A}=\mathrm{Filter}(\mathrm{Sal})$ \hfill{\footnotesize // identify salient textual anchors}

\STATE $\bar{\mathbf{H}}=\frac{1}{|\mathcal{L}_{\mathrm{mid}}|}\sum_{\ell\in\mathcal{L}_{\mathrm{mid}}}\mathbf{H}^{(\ell)}$ \hfill{\footnotesize // stabilize cross-modal representations}

\STATE Extract patch tokens $\mathbf{X}=\{\mathbf{x}_j\}_{j=1}^{N}$ and anchor token reps $\{\mathbf{q}_i\}_{i\in\mathcal{A}}$ from $\bar{\mathbf{H}}$.

\STATE Normalize $\hat{\mathbf x}_j=\mathbf x_j/\|\mathbf x_j\|_2, \hat{\mathbf q}_i=\mathbf q_i/\|\mathbf q_i\|_2\;$; initialize $\mathcal{E}\leftarrow\emptyset$.

\FOR{$i\in\mathcal{A}$}
    \STATE $s_{ij}=\cos(\hat{\mathbf q}_i,\hat{\mathbf x}_j),\;
    w_{ij}=\frac{\exp(s_{ij}/\tau)}{\sum_{u=1}^{N}\exp(s_{iu}/\tau)}$ \hfill{\footnotesize // anchor--patch affinity}

    \STATE $S(\mathcal{B}_i)=\max\limits_{p_{u,v}\in\mathcal{B}_i}s_{u,v},\;
    \mathcal{B}^{*}=\mathrm{TopK}(\{S(\mathcal{B}_i)\})$ \hfill{\footnotesize // score blocks and select top-$k$}
    
    \STATE $\mathcal{R}_i=\mathrm{BBox}\!\left(\bigcup_{\mathcal{B}\in\mathcal{B}^{*}}\mathcal{B}\right)$ \hfill{\footnotesize // merge selected blocks into a region}

    \STATE $E_i=\underset{j\in\mathcal{R}_i}{\mathrm{Concat}}\;\mathbf{x}_j$ \hfill{\footnotesize // concatenate all patch embeddings in region $\mathcal{R}_i$}

    \STATE $\mathcal{E}\leftarrow\mathcal{E}\cup\{E_i\}$
\ENDFOR

\RETURN $\mathcal{E}$
\end{algorithmic}
\end{algorithm}

During each rollout, the model first assesses whether supplementary visual information is required at the current reasoning step. When additional visual information is warranted, the previously cached evidence embeddings are injected into the generation process. Otherwise, the model proceeds directly toward producing an answer. This selection decision can be invoked repeatedly across multiple reasoning steps until the model either produces a final answer or reaches a predefined maximum reasoning horizon. Upon termination, a reward is computed from the final output. When the model fails despite utilizing the cached evidence, we interpret this as an indication that the stored embeddings are misaligned or insufficient. In such cases, we re-identify task-relevant regions under the current model parameters, re-extract their region embeddings, and update the evidence cache accordingly. The refreshed embeddings are employed in subsequent rollouts, enabling the evidence to co-evolve with the policy and progressively improve visual grounding throughout training.

\subsection{Self-Guided Visual Evidence Identification}
\label{sec:training_data}

A central challenge in constructing visual evidence lies in determining what to store: the embeddings must capture precisely the visual information that the model would need to revisit during reasoning, without relying on manual annotation or task-specific heuristics. We address this through a two-stage self-guided visual evidence identification pipeline, which is illustrated in Algorithm \ref{alg:evidence_discovery}. First, the model introspects on its own predictive process to surface the most prediction-critical tokens as textual anchors. Subsequently, we ground these anchors onto spatially coherent image regions via cross-modal matching within the model's internal representation space.

\subsubsection{Discovering Textual Anchors via Gradient Saliency}
\label{sec:concept_discovery}

Rather than relying on external concept taggers or handcrafted keyword lists, we derive anchors directly from the model's own predictive dynamics. Our primary signal is token-level gradient saliency: if a token is critical to the model's next-step prediction, the output logit will exhibit high sensitivity to perturbations of that token's embedding, manifesting as a large gradient magnitude. This yields an importance landscape over input tokens, from which we select the most influential ones as anchors. Formally, let the multimodal model produce logits $\mathbf{z}_L \in \mathbb{R}^{|\mathcal{V}|}$ at the last input position $L$, where $\mathcal{V}$ is the vocabulary. Let $\hat{v}=\arg\max_{v\in\mathcal{V}}\mathbf{z}_L[v]$ denote the predicted next token, and define the scalar target $s=\mathbf{z}_L[\hat{v}]$. We compute a saliency score for each input token embedding $\mathbf{h}_i\in\mathbb{R}^d$ as
\begin{equation}
\mathrm{Sal}(i) = \left\| \nabla_{\mathbf{h}_i} s \;\odot\; \mathbf{h}_i \right\|_2 ,
\end{equation}
where $\odot$ denotes element-wise multiplication. This gradient--input formulation captures both the sensitivity of the prediction (via the gradient) and the magnitude of the representation (via $\mathbf{h}_i$), ensuring that high saliency reflects genuine dependence of the model's output on token $i$. Since raw saliency scores often assign non-trivial weight to function words (eg., \emph{the}, \emph{is}) that carry limited semantic content, we apply a stop-word filter and retain only content-bearing tokens whose saliency exceeds a predefined threshold. The surviving tokens constitute our textual anchors, i.e., the semantics that the model implicitly treats as pivotal to its reasoning (e.g., objects, attributes, or spatial relations). These anchors subsequently serve as queries for visual grounding.

\subsubsection{Identifying Visual Evidence with Textual Anchors}
\label{sec:visual_grounding}

Given the textual anchors, we localize their corresponding visual regions by matching the internal hidden representations of text tokens and image patch tokens within the model's joint multimodal space, where both modalities reside in the same representation space and can therefore be directly compared. Our approach operates on intermediate-layer representations, where cross-modal semantics exhibit more explicit alignment.

Let $\mathbf{H}^{(\ell)} \in \mathbb{R}^{L \times d}$ denote the hidden states at layer $\ell$, for $\ell=1,\dots,\mathcal{L}$. Prior work has shown that middle layers tend to capture richer semantic representations than either early or later layers~\cite{skean2024does, skean2025layer}. We accordingly compute a stabilized representation by averaging middle layer's hidden states with: $\bar{\mathbf{H}}=\frac{1}{|\mathcal{L}_{\mathrm{mid}}|}\sum_{\ell\in\mathcal{L}_{\mathrm{mid}}}\mathbf{H}^{(\ell)}.$ Both modalities are obtained from the same $\bar{\mathbf{H}}$ by indexing the corresponding token positions. Let $\mathbf{X}\in\mathbb{R}^{N\times d}$ denote the patch-token representations and $\mathbf{q}\in\mathbb{R}^{d}$ the representation of a textual anchor token.

To ensure robust similarity computation, we apply mean-centering and $\ell_2$ normalization to both the patch tokens and the anchor, yielding normalized vectors $\{\hat{\mathbf{x}}_i\}_{i=1}^{N}$ and $\hat{\mathbf{q}}$. We then compute anchor--patch affinity via cosine similarity: $s_i=\cos(\hat{\mathbf{x}}_i,\hat{\mathbf{q}})\quad i=1,\dots,N,$
and convert the affinities into a temperature-scaled softmax distribution:
\begin{equation}
w_i=\frac{\exp(s_i/\tau)}{\sum_{j=1}^{N}\exp(s_j/\tau)}.
\end{equation}

We leverage this distribution to extract a spatially coherent region. Specifically, we map patch tokens onto the $H\times W$ patch grid and compute each block's score as the maximum similarity within the block: $s_b = \max_{j \in \mathcal{B}_b} w_{ij},$where $\mathcal{B}_b$ denotes the set of patches in block $b$. We then select the top-$K$ scoring blocks $\mathcal{B}_i = \mathrm{TopK}(\{s_b\}),$
and expand each selected block by computing the bounding rectangle of the selected patches. We set k to 1, and further discussion it in the~\ref{hyperanalysis}. We evaluate the similarity of these patches with their corresponding textual anchors and retain the highest-scoring one as the expanded block $\mathcal{R}_i = \mathrm{Expand}(\mathcal{B}_i)$. The embeddings of patches within each region are then aggregated to form a region-level snapshot: $E_i=\underset{j\in\mathcal{R}_i}{\mathrm{Concat}}\;\mathbf{x}_j$. 
The resulting region embeddings are cached as evidence snapshots and stored alongside each training sample. These embeddings constitute reusable visual evidence that can be dynamically inserted into reasoning during the rollout process.

\subsection{Visual-grounded Reinforcement Learning}
\label{sec:embedding_rl}

Existing tool-augmented thinking-with-images methods~\citep{zheng2025deepeyes, hong2025deepeyesv2, zhang2025thyme} enlarge the action space with external tool calls and require image re-encoding at every reasoning step. Since \sys simply reuses embeddings already produced by the vision encoder and projected into text space, it sidesteps these issues entirely and we can formalize the reasoning as shown in \autoref{eq:formalization}, where the policy selects the next action conditioned on the original image and the full interaction history, including both generated text and any previously inserted visual evidence, accumulated up to the current step:
\begin{equation}\label{eq:formalization}
a_t \sim \pi_\theta(\,\cdot \mid s_t),
\qquad
s_t \triangleq I \,\Vert\, (x_1 \Vert E_1)\,\Vert \cdots \Vert\, (x_{t-1} \Vert E_{t-1}).
\end{equation}
Here, $I$ denotes the input image, $x_t$ is the text generated by the model, and $E_t$ is the embeddings of visual evidence inserted at turn $t$ (with $E_t=\varnothing$ when no insertion occurs). Thus, $s_t$ represents the accumulated context: the raw image followed by all preceding textual responses and inserted evidence blocks in temporal order. Conditioned on $s_t$, the policy samples the next action $a_t$, either terminating with a final answer or triggering insertion of the cached visual evidence. The rollout terminates when a final answer is produced or when a predefined maximum number of turns is reached.

\paragraph{Trajectory-level reward design.}
We design a trajectory-level reward function that holistically evaluates the quality of the complete reasoning path. The reward comprises four complementary components, each yielding a binary score of 1 (satisfied) or 0 (violated). Given a trajectory $\tau$, the total reward is:
\begin{equation}
    \mathcal{R}(\tau) = \lambda_1 \mathcal{R}_{\text{res}}(\tau) + \lambda_2 \mathcal{R}_{\text{format}}(\tau) + \lambda_3 \mathcal{R}_{\text{emb}}(\tau) + \lambda_4 \mathcal{R}_{\text{act}}(\tau),
\end{equation}
where the $\lambda$ values are scaling coefficients. In our experiments, we set $\lambda_1 = 0.6$, $\lambda_2 = 0.3$, $\lambda_3 = 0.5$, and $\lambda_4 = 0.2$. Each component targets a distinct aspect of the desired behavior:

\begin{itemize}[leftmargin=*]
\item \textbf{Format reward} ($\mathcal{R}_{\text{format}}$) promotes well-structured outputs. For single-turn trajectories, the full reward is granted only if the model produces a valid reasoning chain followed by a final answer. For multi-turn trajectories, obtaining the full reward additionally requires an explicit embedding selection during an intermediate turn. Any structural violation results in a zero format reward.

\item \textbf{Result reward} ($\mathcal{R}_{\text{res}}$) evaluates the correctness of the final answer, serving as the primary learning signal for reasoning quality.
\item \textbf{Embedding reward} ($\mathcal{R}_{\text{emb}}$) is activated exclusively when the model produces a correct final answer and invokes embedding insertion at least once during intermediate reasoning steps. This bonus incentivizes the model to actively leverage visual evidence when it is beneficial for task resolution, rather than bypassing the available evidence.

\item \textbf{Action reward} $\mathcal{R}_{\text{act}}$ improves training stability in two ways: (i) it penalizes overly short reasoning traces that could hack the reward, and (ii) it provides a small positive reward for committing to an action, either retrieving an embedding or producing an answer, which discourages the policy from collapsing into ``non-committal'' outputs that avoid taking actions.
\end{itemize}

\section{Experiments}
\label{sec:experiments}


\subsection{Experimental Setup}

\subsubsection{Benchmarks and Baselines}
For training, we sample 1,500 images from COCO 2017~\cite{lin2014microsoft} and construct the corresponding training data with pre-extracted region embeddings. For evaluation, we focus on two challenging high-resolution visual reasoning benchmarks: V*Bench~\cite{wu2024v} and HR-Bench~\cite{hrbench}, reporting results at both 4K and 8K resolutions. To assess generalization beyond high-resolution reasoning, we additionally evaluate on perception benchmarks MME-Real-Lite~\cite{zhang2024mme} and RealWorldQ~\cite{xai_grok15v_2024}; multimodal reasoning benchmarks MathVista~\cite{lu2024mathvista}, LogicVista~\cite{xiao2024logicvistamultimodalllmlogical}, and WeMath~\cite{qiao2025we}; and the hallucination benchmark Hallusion~\cite{wu2024autohallusion}. We compare \sys against representative zoom-and-refine baselines, including DyFo~\cite{li2025dyfotrainingfreedynamicfocus}, ZoomEye~\cite{shen2024zoomeye}, and Zoom-Refine~\cite{yu2025zoom}, as well as a vanilla GRPO-trained model optimized solely with format and accuracy rewards.

\subsubsection{Training Details}
We adopt Qwen3-VL-4B-Instruct~\cite{yang2025qwen3} and Qwen3-VL-8B-Instruct as base models. Both are trained with GRPO~\cite{shao2024deepseekmath} for 60 rollout steps on two NVIDIA H200 GPUs. Each rollout batch contains 16 prompts with 8 rollouts per prompt. We set the KL divergence coefficient to 0.0 and the maximum response length is set as 8,192 tokens.

\subsection{Main Results}

\begin{table*}[t]
\centering
\footnotesize
\setlength{\tabcolsep}{3pt}
\renewcommand{\arraystretch}{1.1}
\resizebox{\textwidth}{!}{
\begin{tabular}{l ccc|ccc|ccc}
\toprule
\multirow{2}{*}{Model}
& \multicolumn{3}{c}{V* Bench}
& \multicolumn{3}{c}{HR-Bench 4K}
& \multicolumn{3}{c}{HR-Bench 8K} \\
\cmidrule(lr){2-4} \cmidrule(lr){5-7} \cmidrule(lr){8-10}
& Attribute & Spatial & Overall
& FSP & FCP & Overall
& FSP & FCP & Overall \\
\midrule

\textit{Qwen3-VL-4B-Instruct} \\

Vanilla$^*$
& 74.78  & 82.89  & 78.01
& 86.00  & 69.50  & 77.75
& 80.75  & 64.00  & 72.38  \\

DyFo
& \underline{86.09}  & 75.00  & 81.68
& 72.75  & 57.25  & 65.00
& 69.75 & 53.50 & 61.62  \\

ZoomEye
& \textbf{92.17}  & 86.84  & \textbf{90.05}
& \textbf{90.00}  & 61.00  & 75.50
& \textbf{88.00} & 60.00 & 74.00  \\

Zoom-Refine
& \underline{86.09}  & 73.47  & 81.17
& 87.00 & 66.00 & 76.50
& \underline{87.00}  & 63.00  & 75.00  \\

GRPO
& 81.74 & \underline{90.79} & 85.34
& \underline{88.50} & \underline{72.50}  & \underline{80.50}
& 83.00 & \textbf{67.75} & \underline{75.38}  \\

\rowcolor{gray!20}
\textbf{\sys}
& 81.74  & \textbf{92.11}  & \underline{85.86}
& 88.00  & \textbf{74.50}  & \textbf{81.25}
& 85.00  & \underline{67.25}  & \textbf{76.13}  \\

\rowcolor{gray!20}
$\Delta$ (vs. Vanilla) 
& +6.96  & +9.22  & +7.85
& +2.00  & +5.00  & +3.50
& +4.25  & +3.25  & +3.75
\\

\midrule
\textit{Qwen3-VL-8B-Instruct}  \\

Vanilla$^*$
& 84.35  & \underline{80.26}  & 82.72
& 89.00  & 69.00  & 79.00
& 81.00  & 67.50  & 74.25  \\

DyFo
& 86.09  & 78.95  & 83.25
& 78.50 & 57.25  & 67.88
& 71.50 & 54.50 & 63.00  \\

ZoomEye
& 81.74  & 73.68  & 78.35
& 88.00  & 60.00  & 74.00 
& \underline{88.00}  & 53.00  & 70.50  \\

Zoom-Refine
& 83.48  & 82.89  & 83.25
& \underline{91.00} & 63.00 & 77.00
& \textbf{92.00}  & 60.00  & 76.00  \\

GRPO
& \underline{86.96}  & \textbf{86.84}  & \underline{86.91}
& \textbf{91.75} & \underline{70.00} & \underline{80.88}
& 86.75 & \underline{68.25} & \underline{77.50}  \\

\rowcolor{gray!20}
\textbf{\sys}
& \textbf{88.70}  & \textbf{86.84}  & \textbf{87.96}
& 90.00  & \textbf{73.00}  & \textbf{81.50}
& 83.00  & \textbf{73.50}  & \textbf{78.25}  \\

\rowcolor{gray!20}
$\Delta$ (vs. Vanilla) 
& +4.35  & +6.58  & +5.24
& +1.00 & +2.75 & +2.50
& +2.00  & +6.00  & +4.00
\\
\bottomrule
\end{tabular}

}
\vspace{3pt}
\caption{Results on V* Bench and HR-Bench. $^*$ indicates results reproduced by our implementation. }
\vspace{-15pt}
\label{tab:main_results}
\end{table*}
In~\autoref{tab:main_results}, we report the performance of \sys~on V* and HRBench, comparing it with other models and methods at both the 4B and 8B scales. As shown in the table, \sys~consistently outperforms all baselines on both benchmarks across both model sizes. Notably, the 4B variant of \sys~achieves a 10.06\% improvement over the corresponding vanilla model. This result indicates that enabling the model to reason with hidden-state embeddings can effectively enhance performance on high-resolution tasks. We further validate our approach on additional datasets. As presented in~\autoref{tab:exp2}, \sys~demonstrates consistent improvements over both the vanilla model and the GRPO-trained model across perception tasks, reasoning tasks, and hallucination benchmarks. In particular, \sys~4B achieves a 17.58\% gain on MME-Real-Lite, while \sys~8B achieves improvements of 19.30\% on MME-Real-Lite and 20.65\% on WeMath. These results suggest that injecting hidden-state embeddings during reasoning can yield benefits across diverse task categories.

\begin{table*}[t]
\centering
\footnotesize
\setlength{\tabcolsep}{4pt}
\renewcommand{\arraystretch}{1.1}
\resizebox{\textwidth}{!}{
\begin{tabular}{c c | c c c | c c c}
\toprule
\multirow{2}{*}{Benchmark} & \multirow{2}{*}{Split}
& \multicolumn{3}{c|}{Qwen3-VL-4B-Instruct}
& \multicolumn{3}{c}{Qwen3-VL-8B-Instruct} \\
\cmidrule(lr){3-5}\cmidrule(lr){6-8}
& 
& Vanilla$^*$ & +GRPO & \sys
& Vanilla$^*$ & +GRPO & \sys \\
\midrule

\multirow{3}{*}{MME-Real-Lite}
& Perception & 50.90 & 54.41 & \textbf{55.95}$_{+5.05}$ 
            & 52.95 & 54.58 & \textbf{58.43}$_{+5.48}$ \\
& Reasoning  & 38.27 & 40.00 & \textbf{51.07}$_{+12.8}$ 
            & 38.40 & 39.47 & \textbf{53.20}$_{+14.8}$ \\
& Overall    & 45.96 & 48.78 & \textbf{54.04}$_{+8.08}$ 
            & 47.26 & 48.67 & \textbf{56.38}$_{+9.12}$ \\
\cmidrule(lr){2-8}

RealWorldQA
& Overall    & 68.10 & 67.84 & \textbf{71.11}$_{+3.01}$
            & 69.28 & \textbf{70.72} & 69.28$_{+0.00}$ \\
\cmidrule(lr){2-8}

MathVista
& Mini       & 67.70 & 68.40 & \textbf{70.50}$_{+2.80}$
            & 69.90 & 70.30 & \textbf{71.60}$_{+1.70}$ \\
\cmidrule(lr){2-8}

LogicVista
& Overall    & 43.08 & 46.21 & \textbf{48.99}$_{+5.91}$
            & 48.66 & 52.01 & \textbf{53.02}$_{+4.36}$ \\
\cmidrule(lr){2-8}

WeMath
& Overall    & 50.98 & 51.03 & \textbf{53.05}$_{+2.07}$
            & 54.71 & 55.40 & \textbf{66.01}$_{+11.3}$ \\
\cmidrule(lr){2-8}

\multirow{4}{*}{HallusionBench}
& qAcc       & 35.74 & 37.55 & \textbf{38.99}$_{+3.25}$
            & 36.10 & 37.91 & \textbf{42.24}$_{+6.14}$ \\
& fAcc       & 41.91 & 42.77 & \textbf{43.64}$_{+1.73}$
            & 44.22 & 45.38 & \textbf{47.98}$_{+3.76}$ \\
& aAcc       & 63.41 & \textbf{64.98} & 64.95$_{+1.54}$
            & 66.35 & 67.19 & \textbf{68.11}$_{+1.76}$ \\
& Overall    & 47.02 & 48.43 & \textbf{49.19}$_{+2.17}$
            & 48.89 & 50.16 & \textbf{52.78}$_{+3.89}$ \\
\bottomrule
\end{tabular}}
\vspace{3pt}
\caption{Benchmark comparison on perception, reasoning, and general tasks. For each model size, we report the base model (Vanilla), applying GRPO to the base model (+GRPO), and our method \sys. Subscripts indicate absolute gains over the corresponding Vanilla baseline. $^*$ denotes results reproduced by us.}
\label{tab:exp2}
\vspace{-25pt}
\end{table*}

\subsection{Visualization and Analysis}

\begin{figure}[htbp]
    \centering
    \includegraphics[width=\textwidth]{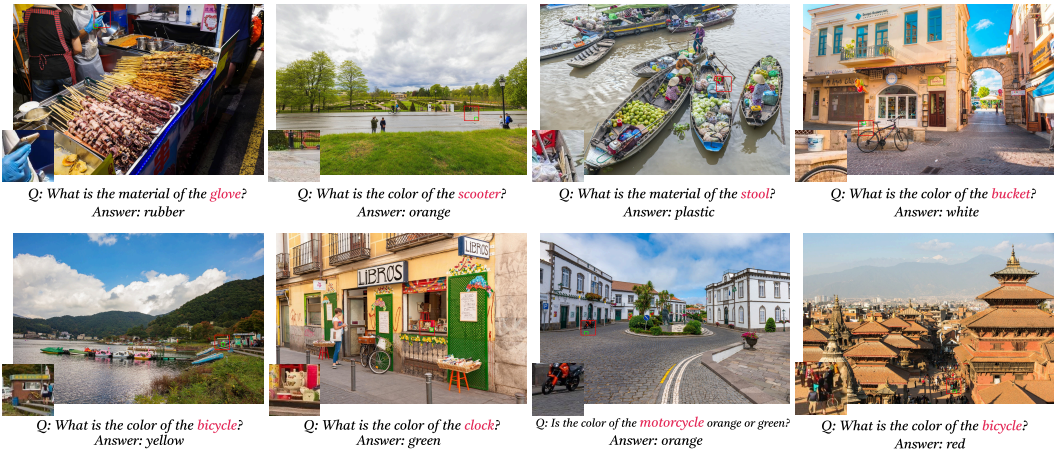}
    \caption{Examples of image regions associated with the extracted embeddings. The \textcolor{darkgreen}{green box} denotes the matched bounding box, and the \textcolor{darkred}{red box} is the expanded region whose embedding is fed into the model. The bottom-left corner shows a zoomed view of the red box. Minor box drift may occur due to Qwen-VL’s patch segmentation.}
    \label{fig:visual_ex}
\end{figure}

In~\autoref{fig:visual_ex}, using the V* dataset as a case study, we demonstrate how our model retrieves bounding boxes in image coordinate space that align with the learned region embeddings. Specifically, the selected embeddings are mapped back to their corresponding spatial patch locations and aggregated to form coherent bounding regions. The resulting visualizations show that these extracted embeddings consistently correspond to semantically meaningful and task-relevant image regions, rather than arbitrary or background areas. Although minor localization drift may occur due to the patch-based segmentation mechanism in Qwen-VL where object boundaries may not perfectly align with fixed patch grids our extended patching strategy mitigates this issue. By explicitly injecting the target object’s embedding as a structured guidance signal during inference, the model is encouraged to focus on spatially relevant regions and refine its reasoning accordingly. This design not only improves visual grounding fidelity but also provides an intuitive explanation for the consistent performance gains achieved by our method across benchmarks.

\subsection{Ablation Studies}






\begin{figure}[t]
    \centering
    \includegraphics[width=\linewidth]{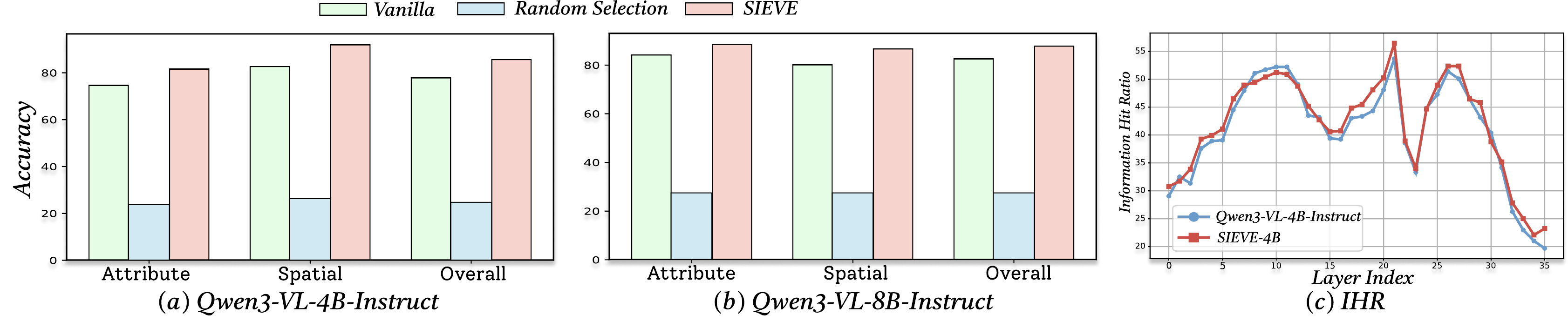}
    \caption{(a) and (b) are the comparison of w/o inserting embedding, inserting random select embedding and insert embedding chosen by our method. (c) is IHR in different Layers on Qwen3-VL-4B-Instruct and \sys-4B.}
    \label{fig:embed}
\end{figure}

\subsubsection{Role of Inserting Embedding}
In this section, we empirically demonstrate that the selected region embeddings contribute positively and meaningfully to the reasoning process. To validate this claim, we construct a controlled ablation experiment in which image patch embeddings are randomly sampled and inserted following the same inference protocol as our method. Concretely, whenever the model determines that additional visual information is required to assist its reasoning, we inject randomly selected patch embeddings instead of the semantically aligned embeddings identified by our saliency-based selection mechanism. We evaluate this variant on V$^*$ Bench, and the results are reported in Figure~\ref{fig:embed}.(a) and Figure~\ref{fig:embed}.(b). As shown, replacing our selected embeddings with randomly sampled ones leads to a substantial and consistent degradation in performance. This performance drop indicates that the gains observed in our method are not simply due to the act of injecting additional visual tokens into the reasoning process. Rather, they stem from incorporating semantically relevant and contextually aligned visual embeddings that meaningfully support intermediate reasoning steps. Together, these results show that our embedding selection captures task-relevant visual information and that the gains stem from informed cross-modal grounding rather than arbitrary token augmentation.

\vspace{-10pt}
\subsubsection{Select Embeddings in Different Layers}
\begin{figure}[t]
    \centering
    \includegraphics[width=\textwidth]{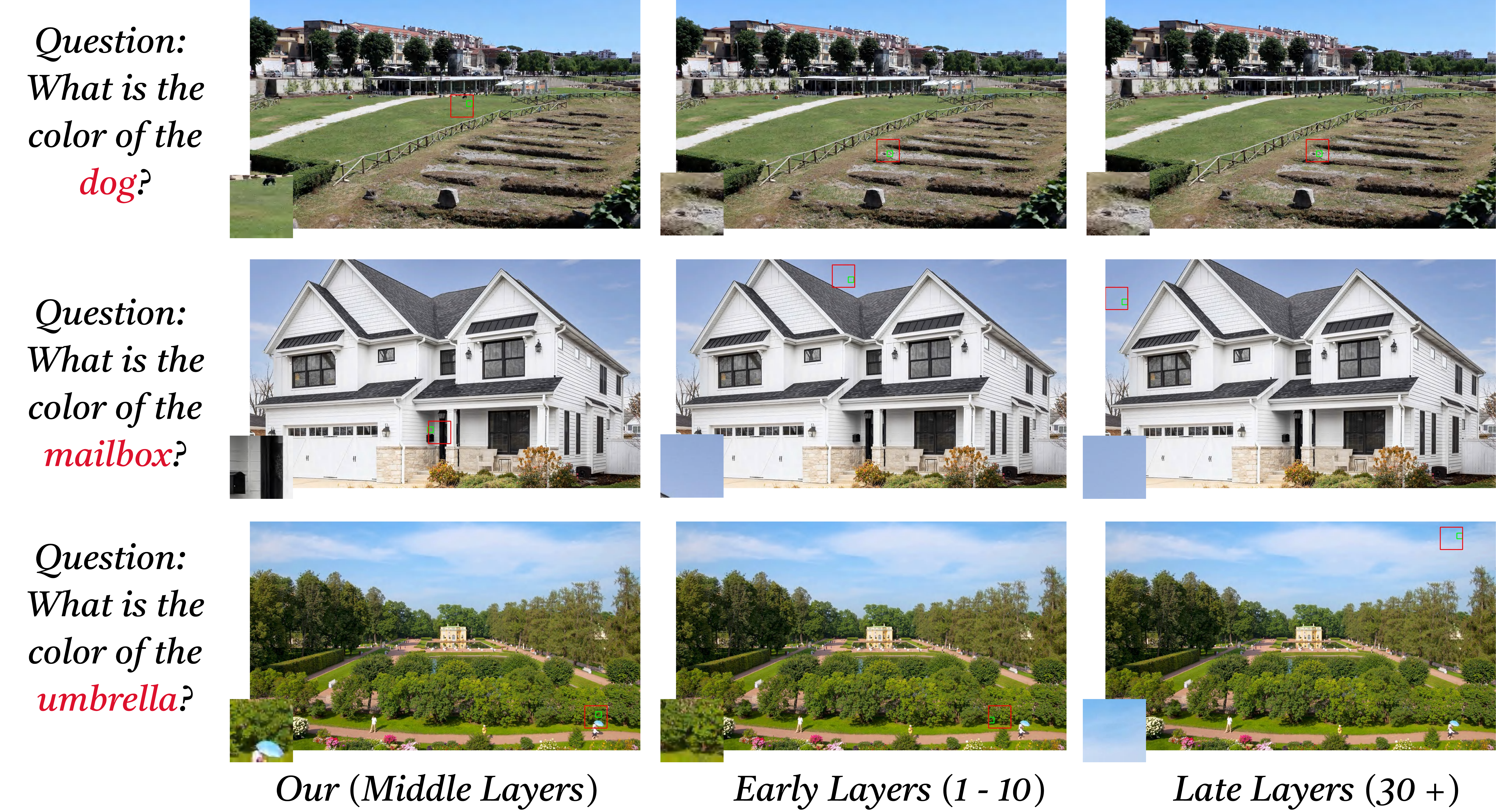}
    \caption{This diagram shows how saliency-guided embeddings for image tokens are computed across transformer layers, highlighting how key regions are identified and how layer choice affects visual grounding quality.}
    \label{fig:visual_layer}
    \vspace{-15pt}
\end{figure}

We visualize image-token hidden states from different VLM layers and analyze the corresponding image embeddings selected via saliency-based token analysis. Specifically, after identifying salient tokens that strongly influence generation, we compute their similarity with image patch representations extracted from different transformer layers and map the selected embeddings back to their spatial locations to obtain visual grounding results.

Using \textit{Qwen3-VL-4B-Instruct} as an example, we compute layer-wise embedding retrieval accuracy to justify selecting middle layers. Since our goal is to extract informative key-region embeddings rather than fully reconstruct object extents, we only require the predicted region to overlap with the ground-truth object region. Accordingly, we define the \textbf{Information Hit Ratio (IHR)} as $\mathrm{IHR} = \mathbb{I}\left( |B_{\mathrm{pred}} \cap B_{\mathrm{gt}}| > 0 \right),$
where a prediction is considered correct if the predicted bounding box overlaps with the ground-truth region.

We randomly sample 300 images from \textit{COCO2017}, each potentially containing multiple objects. As shown in~\autoref{fig:embed}(c), the IHR increases significantly in the middle layers, supporting our design choice of extracting features from intermediate representations. In this experiment, we set $TopK=1$ to directly reflect the feature matching quality of each layer. We further compare the matching precision between \sys and the vanilla model before and after training, showing that \sys achieves higher precision in locating the corresponding object regions. As illustrated in~\autoref{fig:visual_layer}, grounding quality varies across layers. Early-layer representations often fail to capture the true semantic target of the salient token, producing noisy matches, while late-layer representations tend to be overly task-specialized and biased toward output prediction. In contrast, middle-layer representations yield more accurate correspondences between salient tokens and relevant image regions, producing spatially coherent and semantically aligned patches. This observation aligns with prior findings~\cite{skean2024does,skean2025layer} that intermediate transformer layers capture richer semantic information than both shallow and late layers. These results empirically justify computing similarity and extracting region embeddings from middle-layer hidden states, which better balance semantic abstraction and spatial fidelity for cross-modal alignment..

\subsubsection{Role of Action Rewards}
\begin{figure}[t]
    \centering
    \includegraphics[width=\textwidth]{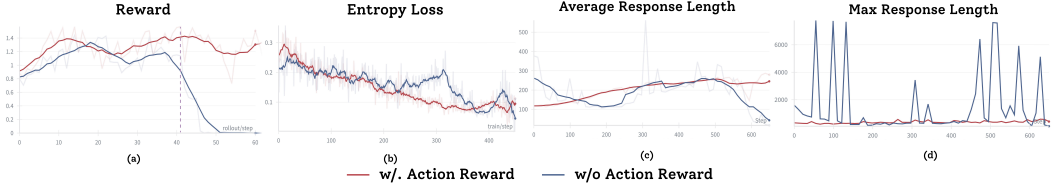}
    \caption{The comparison of metrics change during stage with and without action rewards. (a) is the rewards change. (b) is the extropy loss change, (c) is the average response length, (d) is the max response length.}
    \label{fig:action}
    \vspace{-15pt}
\end{figure}
In Section \ref{sec:embedding_rl}, we incorporate action-level rewards into the total reward function, including a thought richness reward and a signal reward. These additional rewards encourage the model to produce more informative reasoning traces and to issue appropriate response requests, thereby improving both stability and interpretability during training. In~\autoref{fig:action}, we analyze the impact of enabling or disabling these action rewards. Figures (a) and (b) illustrate the effect of the signal reward. In Figure (a), when the signal reward is removed, the training reward collapses in the later stages. Although the model may perform well during early training, the overall reward eventually drops to nearly zero. This indicates that without explicit signal supervision, the policy drifts and fails to maintain consistent behavior. In contrast, Figure (b) demonstrates that introducing the signal reward significantly stabilizes the training process and prevents reward collapse. Figures (c) and (d) examine the influence of the thought richness reward. With this reward enabled, the average response length becomes more stable throughout training, and we no longer observe the sudden and sharp decline in response length that appears when it is disabled. Without the thought richness reward, the model tends to exploit the reward mechanism by outputting empty reasoning tags (e.g., \textit{<think> </think>}) without providing substantive thought content. This shortcut allows the model to obtain rewards without genuinely engaging in reasoning. Furthermore, in the absence of this reward, the maximum response length becomes highly unstable, often leading to repetitive or duplicated content, which artificially inflates the response length. These results show that action-level rewards enhance training stability, prevent degenerate behaviors, and promote meaningful, structured reasoning during reinforcement learning.

\section{Conclusion}
In this work, we propose a training framework, \sys, that enables vision–language models (VLMs) to revisit and leverage image information during inference without relying on external tools. Unlike existing approaches that depend on tool invocation, retrieval systems, or additional visual processing modules, \sys extracts and utilizes the intrinsic signals already present within the VLM’s hidden states, using them as structured hints to guide image-grounded reasoning. By exploiting these internal representations, \sys encourages the model to dynamically refine its understanding of visual content in a lightweight and self-contained manner. Despite its simplicity, \sys demonstrates substantial performance improvements over both the vanilla baseline model and training-free tool-based methods. These results suggest that effective visual revisiting does not necessarily require external tool orchestration; rather, carefully harnessing internal multimodal representations can provide a more efficient and scalable solution for improving image-grounded reasoning.

\bibliographystyle{unsrtnat}
\bibliography{main}
\newpage
\appendix

\section{Time Cost}
\begin{table*}[htbp]
\centering
\caption{Average inference time per item and accuracy on V* benchmark under different reasoning mechanisms.}
\label{tab:time}
\footnotesize
\setlength{\tabcolsep}{4pt}
\renewcommand{\arraystretch}{1.15}
\resizebox{\textwidth}{!}{
\begin{tabular}{l|cc|cc|cc|cc}
\toprule
\multirow{2}{*}{Model} 
& \multicolumn{2}{c|}{Vanilla}
& \multicolumn{2}{c|}{ZoomEyes}
& \multicolumn{2}{c|}{ZoomRefine}
& \multicolumn{2}{c}{\sys} \\
& Time (s) & Acc. 
& Time (s) & Acc. 
& Time (s) & Acc. 
& Time (s) & Acc. \\
\midrule
Qwen3-VL-4B-Instruct & 5.4758 & 78.01 & 5.7689 & 90.05 & 6.5733 & 81.17 & 6.2811 & 85.86 \\
Qwen3-VL-8B-Instruct & 3.5696 & 82.72 & 6.8715 & 78.35 & 7.4327 & 83.25 & 6.4445 & 87.86 \\
\bottomrule
\end{tabular}
}
\vspace{3pt}
\end{table*}

\noindent This section evaluates the relationship between the average response time per question and the accuracy. We run experiments on the Vstar dataset to measure the average time required to answer each question. To ensure a fair comparison across methods, we disable inference acceleration frameworks (e.g., vLLM and SGLang). Accordingly, we exclude DyFo from this analysis, since its original implementation is vLLM-based, and re-implementing it in a huggingface inference stack could introduce non-trivial performance degradation and confound the latency comparison.
All experiments are conducted under the same hardware and runtime conditions. In addition, we fix the maximum generation length to 1024 tokens and enforce a unified “think-then-answer” protocol for all methods, ensuring that differences in response time are not caused by variations in output length or reasoning format. The results are presented in~\autoref{tab:time}. 

Overall, we observe that the average response time of SIEVE does not increase substantially compared to the baseline. This indicates that the additional reasoning mechanism introduced by \sys does not impose significant computational overhead. 
Specifically, compared with methods that require tool invocation during inference, on Qwen3-VL-4B-Instruct, \sys is faster than ZoomRefine, while on Qwen3-VL-8B-Instrct it is fastest compared to tool-call methods. We attribute these differences to the varying reasoning capabilities of different models, which lead to different response behaviors across tasks and consequently different time overheads when combined with various baselines. Overall, although \sys introduces some additional time cost, its overhead is not the largest among the compared methods, while it consistently maintains stable performance improvements across different models and baselines. Considering both time cost and performance, \sys provides the best trade-off between efficiency and effectiveness.

\section{Hyperparameter Analysis}
\label{hyperanalysis}

\begin{figure}[t]
    \centering
    \includegraphics[width=0.5\linewidth]{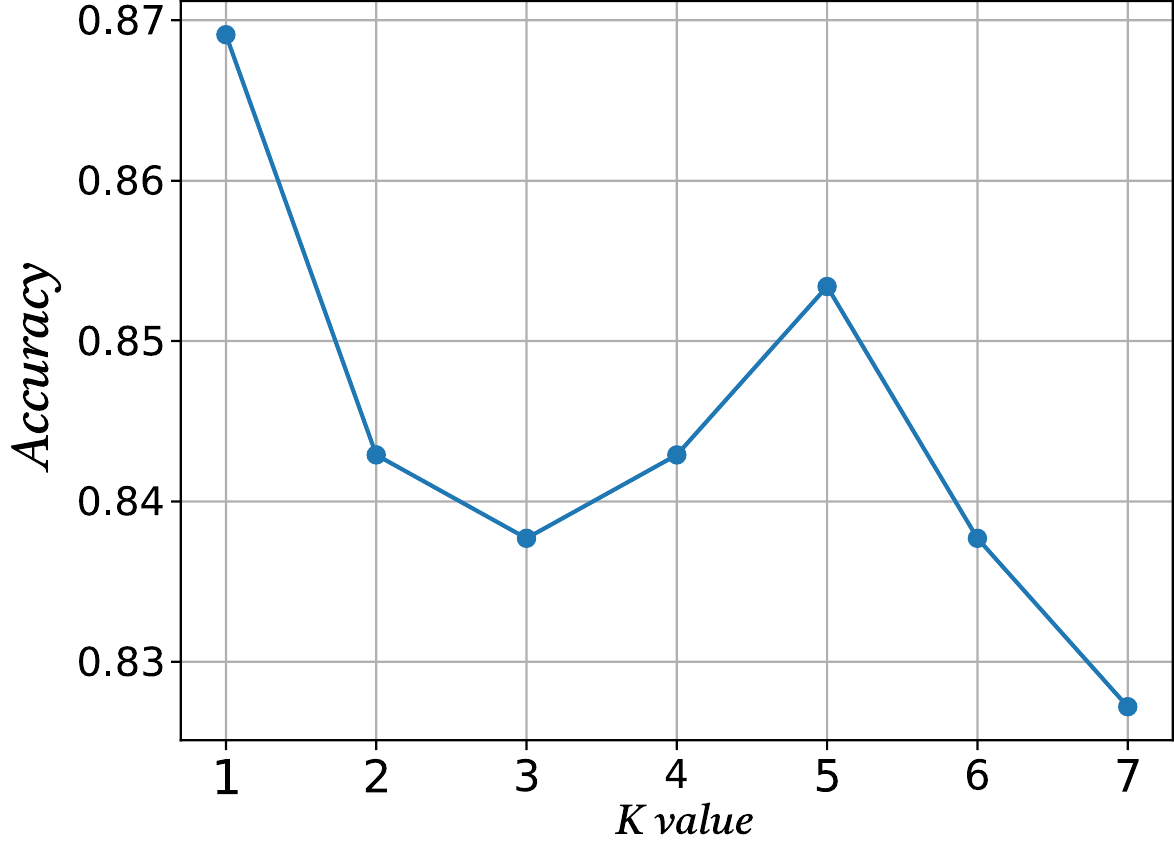}
    \caption{The relationship between the K-value and the accuracy of Vstar benchmark on Qwen3-VL-4B-Instruct.}
    \label{fig:kvalue}
\end{figure}

In this section, we analyze the choice of the top-$k$ parameter in \textit{Identifying Visual Evidence with Textual Anchors} to justify the effectiveness of setting $k=1$. Specifically, we conduct experiments using Qwen3-VL-4B-Instruct on the V* benchmark, varying $k$ from 1 to 7 to evaluate how inserting different numbers of region embeddings affects model performance. The results are shown in~\autoref{fig:kvalue}.

In the Figure, the model achieves the best performance when $k=1$. As $k$ increases, although more embeddings are inserted into the reasoning process, the performance gradually degrades. We attribute this phenomenon to the way region snapshots are constructed. In our method, patches associated with each textual anchor are aggregated to form a single region-level embedding. While this aggregation captures the key visual evidence, it inevitably introduces additional irrelevant visual information within the region. When more regions are inserted (i.e., larger $k$), the amount of such noisy information accumulates, which interferes with the model's reasoning process and reduces performance. These results suggest that inserting a small number of highly relevant visual embeddings is more beneficial than introducing multiple potentially noisy regions. Therefore, we adopt $k=1$ as the default setting in our method.

\section{More Examples}
\begin{figure}[t]
\centering
\includegraphics[width=0.8\linewidth]{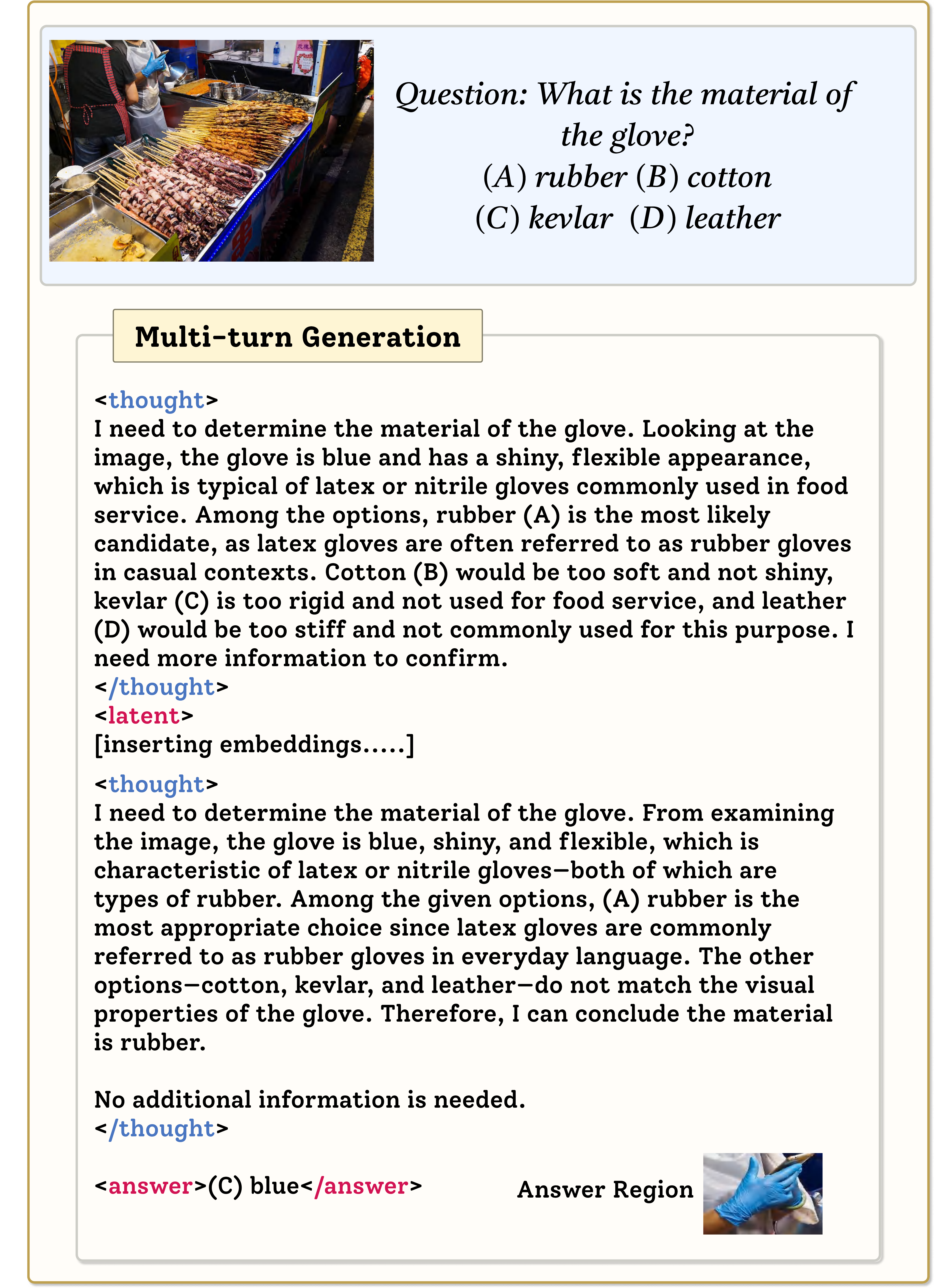}
\caption{Example of \sys on Vstar.}
\label{fig:exa1}
\end{figure}

\begin{figure}[t]
\centering
\includegraphics[width=0.8\linewidth]{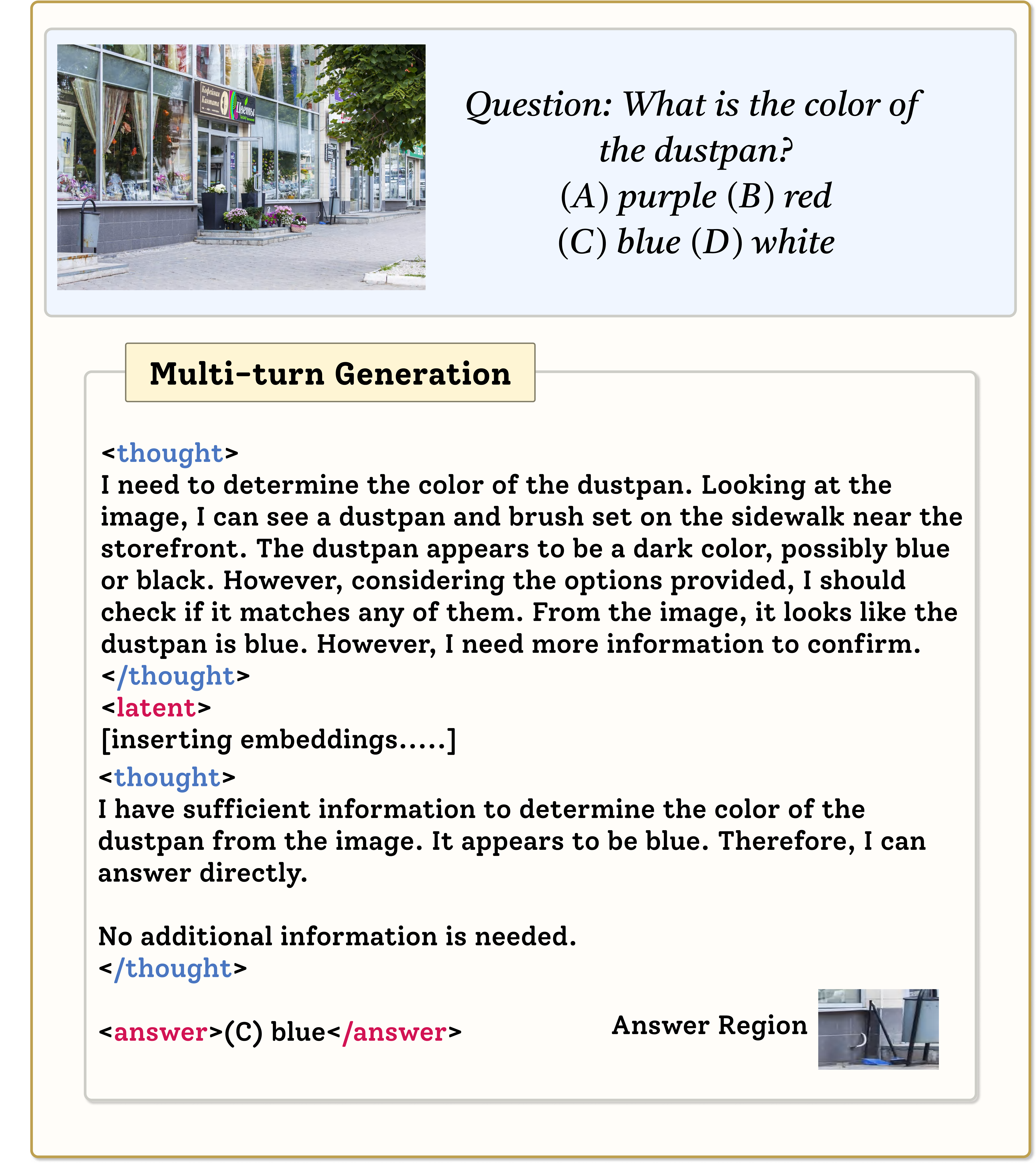}
\caption{Example of \sys on Vstar.}
\label{fig:exa2}
\end{figure}

\begin{figure}[t]
\centering
\includegraphics[width=0.8\linewidth]{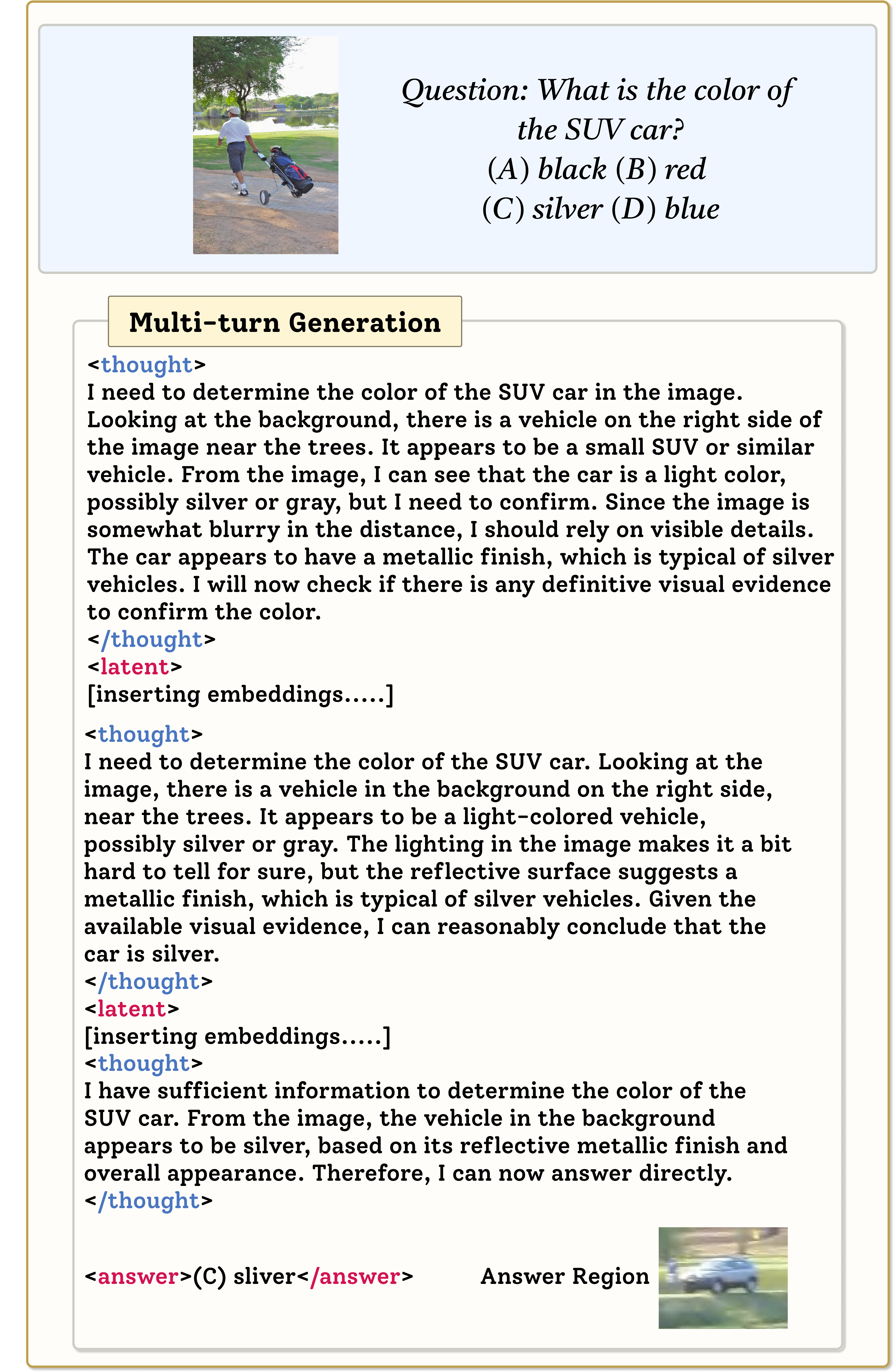}
\caption{Example of \sys on Vstar.}
\label{fig:exa3}
\end{figure}

\begin{figure}[t]
\centering
\includegraphics[width=0.8\linewidth]{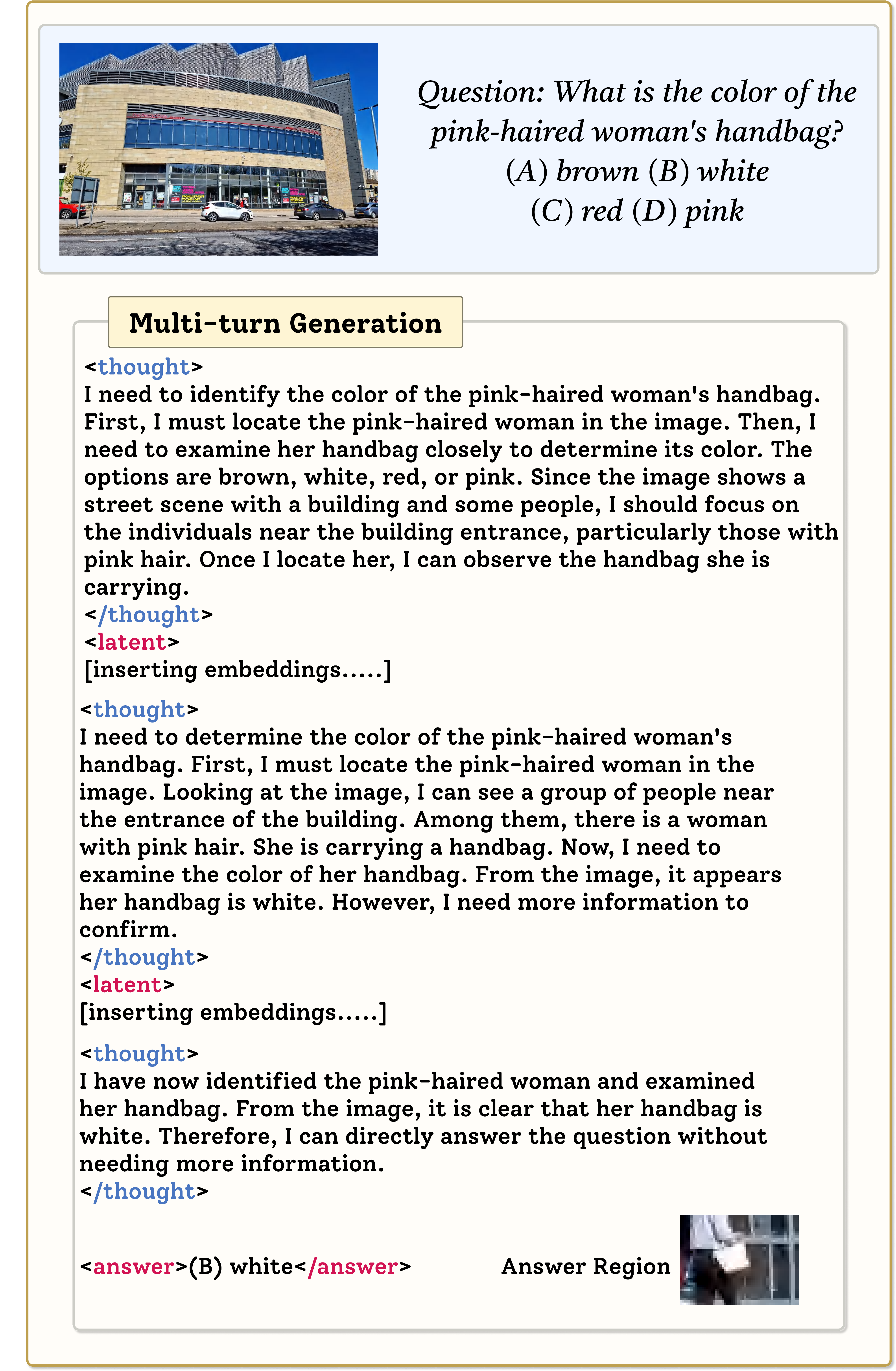}
\caption{Example of \sys on Vstar.}
\label{fig:exa4}
\end{figure}

Here we choose more examples in Vstar benchmark to show the generation process of \sys. In these examples, the content in `[ ]` is not generated by model. We use it to hint of the inserting embeddings operation. We show the example in~\autoref{fig:exa1},~\autoref{fig:exa2},~\autoref{fig:exa3} and~\autoref{fig:exa4}.              

\end{document}